\documentclass[10pt,twocolumn,letterpaper]{article}

\usepackage[pagenumbers]{cvpr} 

\usepackage{graphicx}
\usepackage{amsmath}
\usepackage{amssymb}
\usepackage{booktabs}
\usepackage{enumitem}
\usepackage{multirow}
\usepackage[pagebackref,breaklinks,colorlinks]{hyperref}
\usepackage{algorithmicx,algorithm}
\usepackage{color, colortbl}
\usepackage[dvipsnames]{xcolor}
\definecolor{DnCBG}{rgb}{0.9, 0.9, 1.}

\definecolor{Gray}{gray}{0.5}
\definecolor{GrayBG}{gray}{0.95}
\definecolor{BlueBG}{rgb}{0.9, 0.9, 1.}

\definecolor{nicergreen}{rgb}{0.13, 0.54, 0.13}
\definecolor{nicered}{rgb}{0.83, 0.16, 0.16}
\usepackage[capitalize]{cleveref}
\crefname{section}{Sec.}{Secs.}
\Crefname{section}{Section}{Sections}
\Crefname{table}{Table}{Tables}
\crefname{table}{Tab.}{Tabs.}

\def\cvprPaperID{****} 
\def\confName{CVPR}
\def\confYear{2022}

\newcommand{\gray}[1]{\textcolor{gray}{#1}}
\newcommand{\green}[1]{\textcolor[RGB]{96,177,87}{#1}}
\newcommand{\fn}[1]{\footnotesize{#1}}
\newlength\savewidth\newcommand\shline{\noalign{\global\savewidth\arrayrulewidth\global\arrayrulewidth 1pt}\hline\noalign{\global\arrayrulewidth\savewidth}}
\newcommand{\gbf}[1]{\green{\bf{\fn{(#1)}}}}
\newcommand{\rbf}[1]{\gray{\bf{\fn{(#1)}}}}
\begin{document}
\title{UniVIP: A Unified Framework for Self-Supervised Visual Pre-training}
\author{Zhaowen Li $^{1,2}$ \quad Yousong Zhu $^{1}$ \quad Fan Yang$^{3}$ \quad Wei Li$^{3}$ \quad Chaoyang Zhao $^{1,4}$ \quad Yingying Chen $^{1}$\\ Zhiyang Chen $^{1,2}$ \quad Jiahao Xie $^{5}$\quad Liwei Wu $^{3}$\quad Rui Zhao $^{3,7}$ \quad Ming Tang $^{1}$\quad Jinqiao Wang $^{1,2,6}$\\
  National Laboratory of Pattern Recognition, Institute of Automation, CAS, Beijing, China$^{1}$ \\ 
  School of Artificial Intelligence, University of Chinese Academy of Sciences, Beijing, China$^{2}$\\
  SenseTime Research$^{3}$ \ Development Research Institute of Guangzhou Smart City, Guangzhou, China$^{4}$ \\ S-Lab, Nanyang Technological University$^{5}$ \ Peng Cheng Laboratory, Shenzhen, China$^{6}$
   \\ Qing Yuan Research Institute, Shanghai Jiao Tong University, Shanghai, China$^{7}$ \\
   \\
{\tt\small \{zhaowen.li,yousong.zhu,chaoyang.zhao,yingying.chen,zhiyang.chen,tangm,jqwang\}@nlpr.ia.ac.cn}\\
{\tt\small  \{yangfan1,liwei1,wuliwei,zhaorui\}@sensetime.com} \quad {\tt\small  jiahao003@ntu.edu.sg}
}
\maketitle
\begin{abstract}
Self-supervised learning (SSL) holds promise in leveraging large amounts of unlabeled data. However, the success of popular SSL methods has limited on single-centric-object images like those in ImageNet and ignores the correlation among the scene and instances, as well as the semantic difference of instances in the scene. To address the above problems, we propose a Unified Self-supervised Visual Pre-training (UniVIP), a novel self-supervised framework to learn versatile visual representations on either single-centric-object or non-iconic dataset. The framework takes into account the representation learning at three levels: 1) the similarity of scene-scene, 2) the correlation of scene-instance, 3) the discrimination of instance-instance. During the learning, we adopt the optimal transport algorithm to automatically measure the discrimination of instances. Massive experiments show that UniVIP pre-trained on non-iconic COCO achieves state-of-the-art transfer performance on a variety of downstream tasks, such as image classification, semi-supervised learning, object detection and segmentation. Furthermore, our method can also exploit single-centric-object dataset such as ImageNet and outperforms BYOL by 2.5\% with the same pre-training epochs in linear probing, and surpass current self-supervised object detection methods on COCO dataset, demonstrating its universality and potential.
\end{abstract}

\section{Introduction}
\label{sec:intro}

\begin{figure}[ht]
  \centering
  \includegraphics[width=0.98\linewidth]{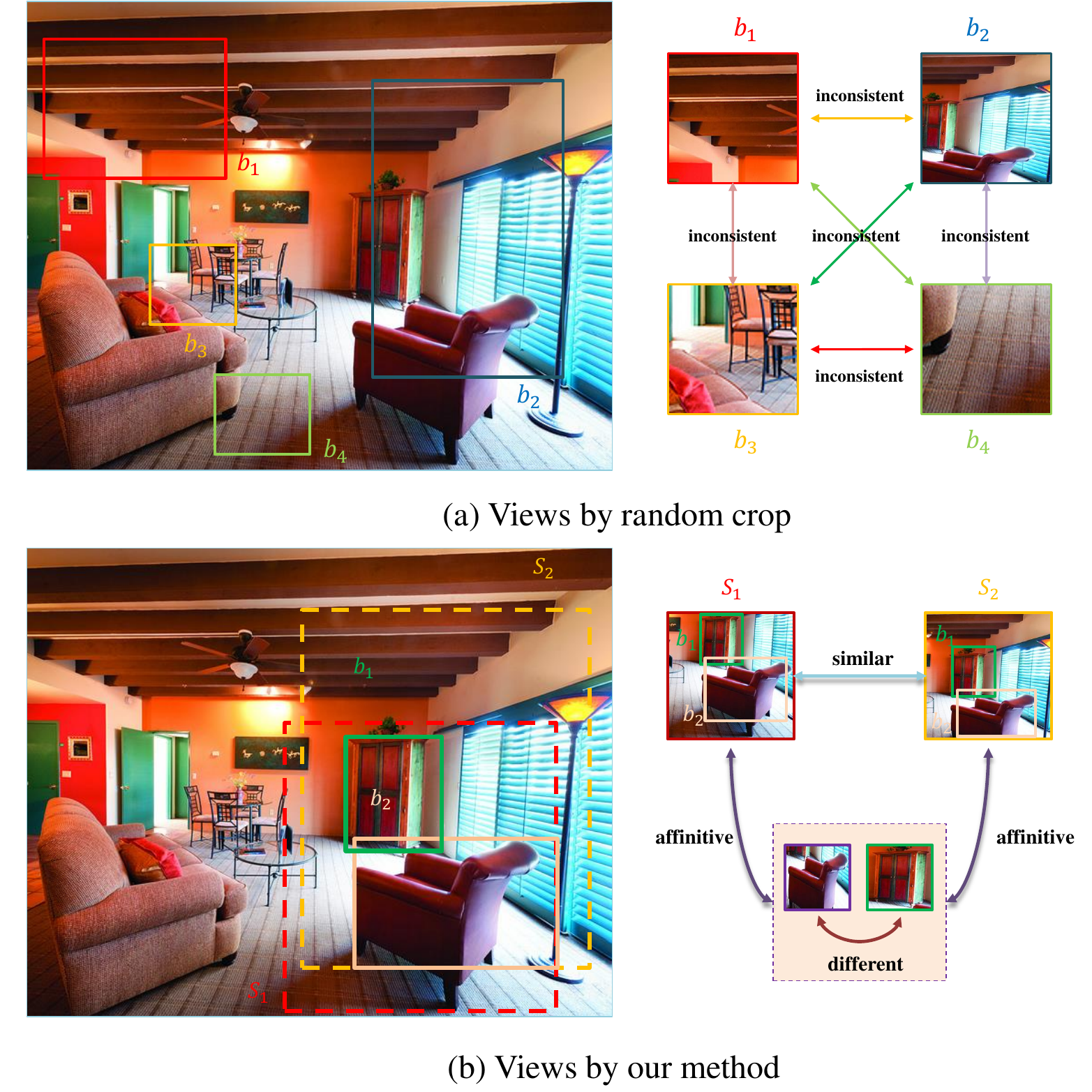}
  \vspace{-2mm}
  \caption{\textbf{Visualization of different cropped views on COCO dataset.} (a) Images from COCO contain multiple instances, thus different random crops might represent different semantic meanings and are not satisfied with the semantic consistency assumption. (b) Our method for unified self-supervised learning. The two scene views are created with overlapping regions while the overlapping regions contain multiple instances. }
  \label{fig1}
  \vspace{-4mm}
\end{figure}

Deep learning has shown excellent performance on various computer vision tasks \cite{deng2009imagenet,he2016deep,ren2015faster,lin2020focal,he2017mask,Yunze2020Progressive} using labels. Self-supervised learning (SSL) of visual representation aims at capturing salient feature representation without relying on human annotations. Recently, contrastive learning \cite{wu2018unsupervised,he2019momentum,chen2020generative,chen2020simple,chen2020improved,chen2020big,grill2020bootstrap,caron2020unsupervised,li2021mst} based SSL has proved impressive results on a number of downstream tasks, largely narrowing the gap between unsupervised and supervised learning and even surpassing the supervised counterpart. These state-of-the-art (SOTA) methods build upon the pretext task called instance discrimination, which regards different views of a single image as the same instance and its objective is simply to learn a feature representation that discriminates among images. Hence, the basic assumption of these methods is that the pre-training data should have the property of semantic consistency \cite{liu2020self,chen2021multisiam}, \ie, the assumption highly relies on the single-centric-object data such as those in ImageNet \cite{deng2009imagenet}. Nevertheless, it is infeasible for complex natural images since they usually consist of multiple instances as shown in Fig~\ref{fig1}(a). Some work \cite{grill2020bootstrap,misra2020self,selvaraju2021casting} naively extend the off-the-shelf SSL methods from ImageNet to other datasets, like MS COCO \cite{lin2014microsoft}, Places365\cite{zhou2017places}, and YFCC100M \cite{thomee2016yfcc100m}, yet they do not acquire satisfactory results. 

It is known that multiple instances in the single natural image possess the co-occurrence relationship, and usually have different semantic meanings. Therefore, the models should have the ability to distinguish the semantics of different instances. However, it is still challenging to discriminate different instances residing in the single natural image when no instance annotations are available. Several region-level based methods \cite{liu2020self,pinheiro2020unsupervised,roh2021spatially} propose to leverage multiple local regions to pre-train models using non-iconic dataset, and achieve the success of the specific downstream task. Nevertheless, these region-level based methods do not explicitly distinguish different instances in the scene. In addition, their results of linear evaluation are inferior to the baseline, \ie, these methods can not obtain versatile visual representations. Moreover, natural images have prior that the scene and instances in the scene have the semantic affinity since these instances correlate with the scene. Current SSL methods are not aware of the prior and do not encode the semantic affinity. Because of the above problems, the application scenario of these methods is limited. It is essential to design an effective learning paradigm to obtain versatile visual representations.

In this paper, we introduce a unified self-supervised pre-training framework, named UniVIP, to learn the visual representations by pre-training on either single-centric-object or non-iconic dataset. Specifically, we first exploit the unsupervised instance proposal method Selective Search \cite{uijlings2013selective} to generate candidate instances. Then, for each image, we create two scene views with overlapping regions containing instances to guarantee the global similarity, \ie, the similarity of scene-scene, as much as possible, which will effectively alleviate the semantic inconsistency of different scene views. Moreover, to tackle the correlation of scene-instance, the generated instances are grouped to approximate the semantics of the corresponding scene views, guiding the network to learn a variety of instances in the image. In our UniVIP, the discrimination of instance-instance is formulated as the optimal matching problem among all candidate instances in overlapping regions and uses the optimal transport algorithm \cite{ge2021ota} to discriminate different instances in the scene. Our objective consists of the above three items, and different views of some scenes and instances obtained by our UniVIP are shown in Fig~\ref{fig1}(b). It is noted that our framework is specially designed to learn versatile representations from natural images, and is able to fully leverage the prior of semantic affinity among the natural scene and instances in the scene, and explicitly distinguish co-occurrence instances.

Massive experiments in single-centric-object and non-iconic datasets prove that UniVIP can learn the versatile representations. In particular, our method outperforms the state-of-the-art by 2.3\% top-1 classification accuracy with pre-training on COCO dataset for the ImageNet \cite{deng2009imagenet} linear evaluation protocol. Our 300-epoch UniVIP achieves 42.2 bbox mAP and 38.2 mask mAP using Mask R-CNN \cite{he2017mask} on COCO detection and segmentation with 1$\times$ schedule when pre-trained on ImageNet, and even surpasses the popular self-supervised object detection methods.

Overall, we make the following contributions:
\begin{itemize}[leftmargin=0.2in]
    \setlength{\itemsep}{0pt}
    \setlength{\parsep}{0pt}
    \setlength{\parskip}{0pt}
    \item We proposed a Unified Self-supervised Representations Learning framework to effectively overcome the semantic inconsistency of random views in non-iconic images, and it can be pre-trained with any images. 
     
    \item We proposed to simultaneously leverage the similarity of scene-scene, the correlation of scene-instance, and the discrimination of instance-instance to promote the performance of models effectively.

    \item Extensive experiments demonstrate the effectiveness and stronger generalization ability of our method. Specifically, the models pre-trained with UniVIP on single-centric-object and non-iconic datasets all outperform previous SOTA methods in multiple downstream tasks, such as image classification, semi-supervised learning, object detection and segmentation.

\end{itemize}

\section{Related work}

\begin{figure*}[ht]
  \vspace{-2mm}
  \centering
  \includegraphics[width=1.0\linewidth]{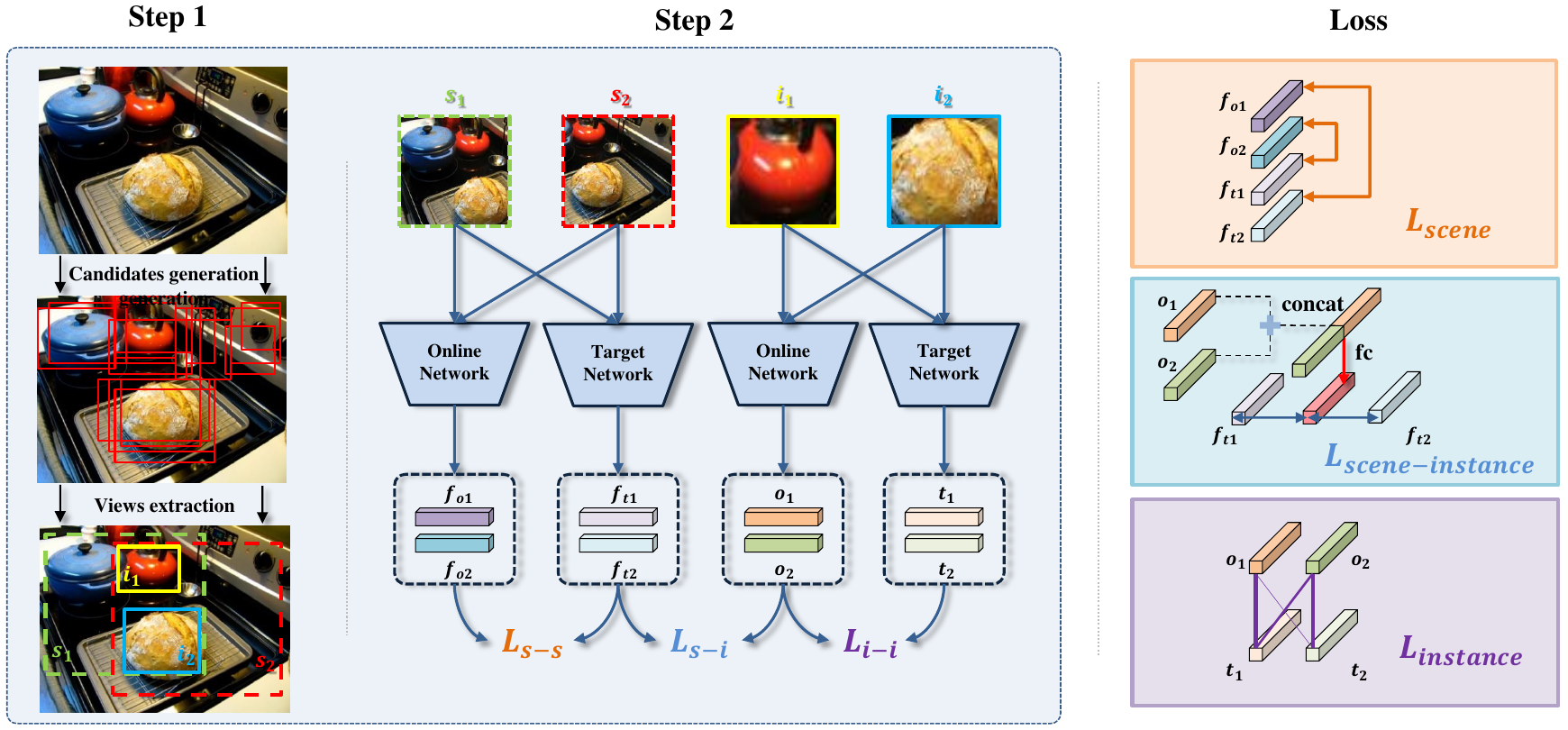}
  \vspace{-6mm}
  \caption{\textbf{The pipeline of UniVIP.} The non-iconic image is first extracted candidate instances using unsupervised object proposal algorithms selective search. Then, we create two views with overlapping regions and multiple instances in the overlapping regions from the image. The existence of overlapping regions can guarantee the scene's similarity. Here, we adopt two instances as an example. Furthermore, we feed the two scene views and two instances to the online and target networks, and obtain feature representations. Finally, we compute the similarity of scene-scene, the correlation of scene-instance, and the discrimination of instance-instance. The total loss function consists of the $\mathcal{L}_{\text{scene}}$, $\mathcal{L}_{\text{scene-instance}}$, and $\mathcal{L}_{\text{instance}}$.} 
  \label{ourmethod}
  \vspace{-6mm}
\end{figure*}

\subsection{SSL on single-centric-object dataset}

Currently, the most competitive pretext task for self-supervised visual representation learning is instance discrimination \cite{he2019momentum,chen2020simple,chen2020improved,chen2020big,grill2020bootstrap,caron2020unsupervised}. The learning objective is simply to learn representations by distinguishing each image from others, and this approach has proved excellent performance on extensive downstream tasks, such as image classification \cite{deng2009imagenet,he2016deep}, object detection \cite{ren2015faster,lin2020focal} and segmentation \cite{he2017mask}. MoCo \cite{he2019momentum} improves the training of instance discrimination methods by storing representations from a momentum encoder instead of the trained network.
SimCLR \cite{chen2020simple} shows that the memory bank can be entirely replaced with the elements from the same batch if the batch is large enough. Meanwhile, BYOL \cite{grill2020bootstrap} directly bootstraps the representations by attracting the different features from the same instance. SwAV \cite{caron2020unsupervised} maps the image features to a set of trainable prototype vectors and proposes a multi-crop strategy for self-supervised data augmentation. Moreover, Some works \cite{chen2021an,caron2021emerging} extend the pretext task to vision Transformer \cite{dosovitskiy2021an,chen2021dpt} and achieve superior performance in image classification. Furthermore, MST \cite{li2021mst} propose an attention-guided mask strategy to avoid masking crucial regions of images for self-supervised Transformer learning.
Nevertheless, much of their progress has far been limited to single-centric-object pre-training data such as ImageNet \cite{deng2009imagenet}, and may be infeasible when extended to non-iconic datasets.

\subsection{SSL on natural scene}

Purushwalkam \textit{et al.} \cite{purushwalkam2020demystifying} point out that the advance of current SSL methods moderately comes from their usage of dataset bias of ImageNet. Also, they find that training MoCo on the less-biased COCO dataset does not get encouraging results. 
Moreover, HED \cite{zhang2020self} report degraded performance when training MoCo on COCO and PASCAL \cite{everingham2010pascal} datasets. 
Recently, MaskCo \cite{zhao2021self} also notices that the semantic consistency assumption of current SSL methods and proposes a contrastive mask prediction task for visual representation learning.
Some preliminary work \cite{grill2020bootstrap,misra2020self,selvaraju2021casting} naively extend the off-the-shelf constrastive learning methods from ImageNet to other datasets, like MS COCO \cite{lin2014microsoft}, Places365\cite{zhou2017places}, and YFCC100M \cite{thomee2016yfcc100m}, yet they do not acquire satisfactory results since these datasets are not satisfied with the semantic consistency assumption, even though the size of some datasets is orders of magnitude larger than ImageNet. Furthermore, DenseCL \cite{pinheiro2020unsupervised}, Self-EMD\cite{liu2020self}, MaskCo and SCLR\cite{roh2021spatially} leverage local regions of non-iconic images to pre-train the models, yet these methods only work on specific downstream task and can not acquire versatile visual representations. DnC \cite{tian2021divide} alternates between contrastive learning and clustering-based hard negative mining to train YFCC100M \cite{thomee2016yfcc100m} and JFT-300M \cite{sun2017revisiting}. ORL \cite{xie2021unsupervised} shows the impressive performance when pre-trained on MS COCO, but its three-stage method consumes much time, thus cannot support large-scale pre-training. These methods, however, are not aware of the semantic affinity between the scene and instances, and also ignore the semantic discrimination of different instances.

\section{Approach}

The pipeline of our proposed UniVIP is shown in Fig~\ref{ourmethod}. We propose a unified visual self-supervised approach to learn versatile visual representations, which creatively integrates the scene similarity, the scene-instance semantic affinity, and the semantic discrimination of different instances. Here, we first review the basic instance discrimination method in \ref{sec:preliminary}. Then, the mechanism and effect of the scene similarity are explained in \ref{scene}. Furthermore, we study the correlation of scene-instance in \ref{sc-in}. Finally, the optimal transport algorithm is imported to promote the semantic discrimination of different instances, and the training function of our method is described in \ref{discrimination}.

\subsection{Preliminary}
\label{sec:preliminary}

Without loss of generality, we adopt BYOL \cite{grill2020bootstrap} as our basic self-supervised learning method, which can achieve state-of-the-art transfer performance. For each image $x$, BYOL first generates two views $x_1 \sim \mathcal{T}_{1}(x)$ and $x_2 \sim \mathcal{T}_{1}(x)$ under random data augmentation, which are then fed into the \emph{online} network $f_\theta(x)$ and the \emph{target} network $g_\xi(x)$ separately, parameterized by $\theta$ and $\xi$. Both online and target networks possess a neural network backbone and a projection head \cite{chen2020big}, which share the same architecture with different parameters. While the online network has the target predictor \cite{grill2020bootstrap}. The parameters $\xi$ of fixed network $f_\xi(x)$ is updated by the exponential moving average of the parameters $\theta$ of online network according to Eq (\ref{mam}), where decay rate $ m \in [0, 1]$ is the momentum and will increase to 1.0 until the end of training.
\begin{equation}
    \xi = m * \xi + (1 - m) * \theta \label{mam}
\end{equation}

Finally, BYOL maximizes the cosine similarity between the prediction of online network and the projected feature of target network as the scene-level consistency. The loss function is defined as Eq (\ref{cross}).
\begin{equation}
    \mathcal{L}(x_1, x_2) \triangleq -\frac{\langle f_\theta\left(x_1\right), g_\xi\left(x_2\right) \rangle}{||f_\theta\left(x_1\right)||_2 \cdot ||g_\xi\left(x_2\right)||_2},
    \label{cross}
\end{equation}

\subsection{Similarity of scene-scene}
\label{scene}

The semantic consistency assumption is almost always satisfied in the single-centric-object ImageNet, which is the highly curated pre-training dataset. However, the implicit assumption can not be scalable to the natural dataset with non-iconic images. The main reason why inconsistency happens in non-iconic images is that the two random views may be far away from each other.

Meanwhile, the instance annotations of the natural images are unavailable. Therefore, to acquire the candidate instances as the prior, we leverage the unsupervised instance proposal algorithms selective search to generate proposals for each image. To filter the certain number of redundancy of generated proposals, we set some pre-defined thresholds including the minimal scale, the range of aspect ratio, and the maximal intersection-over-union (IoU) among these instance-based regions. Considering the instances that exist in the natural scene, we create two scene views $\bf s_1$, $\bf s_2$ with the overlapping regions containing $K$ identical instances. For ensuring that each image has $K$ regions, we generate boxes by the \emph{naive} strategy if the number of candidate instances in the overlapping regions is less than $K$. The naive strategy includes setting the minimum scale to $64$ pixels, the range of aspect ratio is between $1/3$ and $3/1$, and the maximum IoU threshold is $0.5$.

By constructing two scene views with overlapping regions, we transfer the semantic inconsistency of random views to similarity of scene-scene in natural images. In particular, we feed the two scene views to the online and target network separately, and acquire the representations $\bf{f_{o1}}$, $\bf{f_{o2}}$, $\bf{f_{t1}}$, and $\bf{f_{t2}}$ to compute the symmetric loss as Eq (\ref{loss_scene}), following BYOL \cite{grill2020bootstrap}.
\begin{equation}
    \mathcal{L}_{\text{scene}} =  \mathcal{L}(\mathbf{s_1}, \mathbf{s_2}) + \mathcal{L}(\mathbf{s_2},\mathbf{s_1})
    \label{loss_scene}
\end{equation}

\subsection{Correlation of scene-instance}

\label{sc-in}

Concretely, natural images have prior that the scene and instances residing in it possess the semantic affinity since these instances correlate with the scene. Apparently, it is reasonable to argue that exploring the prior among the scene and the instances is conducive to learning more general feature representations. However, current un-/self-supervised learning methods do not take into account the existence of the correlation of scene-instance. To research the correlation in the un-/self-supervised learning field, the primary issue is how to measure the semantic affinity among the scene and instances. 
Due to the simplicity and effectiveness of the cosine similarity, UniVIP attempts to establish the semantic affinity among the scene and multiple instances by the cosine similarity. Specifically, we crop and resize each instance $i_k$ in overlapping regions to $96 \times 96$, where $k = \{1,...,K\}$, then we feed $K$ instances into the online network and obtain $K$ representations vectors $ [\mathbf{o_1},\mathbf{o_2}, ...,\mathbf{o_K}]$. Furthermore, we concatenate these representations, and linearly map the concatenated representation to the dimension of the scene's representation and obtain the final representation $\bf{I}$ as Eq (\ref{fuse}). 
\begin{equation}
    \mathbf{I} = f_{\text{linear}}(\text{concat}(\mathbf{o_1},\mathbf{o_2},...,\mathbf{o_K}))
    \label{fuse}
\end{equation}

Finally, we minimize the cosine distance of the scene view $\bf s$ and instances combination in the feature space as Eq (\ref{corres}) and argue that the measurement can explore the prior of scene-instance semantic affinity.
\begin{equation}
    \mathcal{L}_{\text{affinity}}(\mathbf{s}, \mathbf{I}) \triangleq -\frac{\langle \mathbf{I}, g_\xi\left( \mathbf{s} \right) \rangle}{|| \mathbf{I}||_2 \cdot ||g_\xi\left(\mathbf{s}\right)||_2}
    \label{corres}
\end{equation}

We also compute the symmetric views since the overlapping area of two views possesses the same instances. Thus, the semantic affinity can be explored accroding to Eq (\ref{s-i}).
\begin{equation}
    \mathcal{L}_{\text{scene-instance}} = \mathcal{L}_{\text{affinity}}(\mathbf{s_1},\mathbf{I}) + \mathcal{L}_{\text{affinity}}(\mathbf{s_2},\mathbf{I})
    \label{s-i}
\end{equation}

\subsection{ Discrimination of instance-instance}
\label{discrimination}

In subsection \ref{sc-in}, we increase the affinity among the scene and instances, yet can not ensure that extracted features in each instance can be distinguished from other instances. Moreover, the contrastive loss \cite{he2019momentum} demands many negative samples, while the number of instances in the non-iconic image is limited, which can not meet the demand. Therefore, in this subsection, we formulate the discrimination of instance-instance as an optimal transport problem. Here, we first describe the concept of optimal transport, then introduce how to apply the optimal transport to train the models for learning visual feature representations. Finally, we establish the training function of UniVIP.

\textbf{Optimal transport.}
The form of Optimal Transport (OT) can be described as the following problem: supposing that a set of $M$ suppliers are required to transport goods to a set of $N$ demanders. 
The $m$-th supplier holds $b_m$ units of goods while the $n$-th demander needs $a_n$ units of goods. The cost per unit transported from supplier $m$ to demander $n$ is denoted by $c_{mn}$. The goal of optimal transport algorithm is to find a transportation plan $\tilde{\mathcal{Y}}=\{y_{m,n} | m=1,2,...M, n=1,2,...N\}$, according to which all goods from suppliers can be transported to demanders at a minimal transportation cost as Eq (\ref{origin_formulation}).
\begin{alignat}{2}
\begin{split}
\min_{y}\quad &\sum\limits_{m=1}^{M}\sum\limits_{n=1}^{N} c_{mn}y_{mn}. \\
\mbox{s.t.}\quad&\sum\limits_{m=1}^{M}y_{mn}=a_n,\quad \sum\limits_{n=1}^{N}y_{mn}=b_m, \\
&\sum\limits_{m=1}^{M}b_m=\sum\limits_{n=1}^{N}a_n, \\
&y_{mn}\geq 0,\quad m=1,2,...M, n=1,2,...N.
\end{split}\label{origin_formulation}
\end{alignat}

\textbf{OT for semantic discrimination.}

For feeding the candidate instances in overlapping regions to the online network, there have a set of feature vectors $[\mathbf{o_1},\mathbf{o_2}, ...,\mathbf{o_K}]$, and each vector $o_m$ can be seen as a node in the set. Moreover, we also feed these instances to the target network. Similarly, the set of feature vectors $ [\mathbf{t_1},\mathbf{t_2}, ...,\mathbf{t_K}]$ can be acquired by the target network. Following the original optimal transport formulation in Eq (\ref{origin_formulation}), the cost per unit transported from supplier feature node $\mathbf{o_m}$ to demander node $\mathbf{t_n}$ is defined as Eq (\ref{cost}). Thus, the nodes with similar representations tend to generate fewer transport cost between each other while the nodes with irrelevant representations tend to generate more transport cost. The similarity of each pair of instances can be represented as the optimal matching cost between two sets of vectors.
\begin{equation}
c_{mn} = 1-\frac{{\mathbf{o}_m}^T  \mathbf{t}_n}{\lVert \mathbf{o}_m\rVert \lVert \mathbf{t}_n\rVert}
\label{cost}
\end{equation} 

Moreover, the marginal weights $a_m$ and $b_n$ are defined as Eq (\ref{weight1}) and Eq (\ref{weight2}), where the function $max(\cdot)$ ensures the weights are always non-negative. 
\begin{equation}
b_{m} = \max\{ \textbf{o}_m^T \cdot \frac{\mathbf{f_{t1}}+\mathbf{f_{t2}}}{2} , 0\} 
\label{weight1}
\end{equation} 
\begin{equation}
a_{n} = \max\{ \textbf{t}_n^T \cdot  \frac{\mathbf{f_{o1}}+\mathbf{f_{o2}}}{2}, 0\}
\label{weight2}
\end{equation} 

Following \cite{ge2021ota}, we address the Eq (\ref{origin_formulation}) by a fast iterative solution, named Sinkhorn-Knopp \cite{cuturi2013sinkhorn}, and acquire the optimal matching flows $\tilde{\mathcal{Y}}$. Then we can compute the discrimination of instance-instance as Eq (\ref{instance}). Here, the loss would be minimized only if the representations of each instance are similar to itself and dissimilar to other instances, \ie, the instance can be distinguished from other instances.

\begin{equation}
    \mathcal{L}_{\text{instance}}(\mathbf{O},\mathbf{T})\triangleq -\sum_{m=1}^{K}\sum_{n=1}^{K}\frac{{\mathbf{o}_m}^T  \mathbf{t}_n}{\lVert \mathbf{o}_m\rVert \lVert \mathbf{t}_n\rVert} \tilde{y}_{mn} \label{instance}
\end{equation}

Finally, the total loss of our UniVIP is formulated as Eq (\ref{final}), and each loss coefficient is equally weighted.
\begin{equation}
    \mathcal{L}_{\textbf{UniVIP}} = \mathcal{L}_{\text{scene}} + \mathcal{L}_{\text{scene-instance}} +  \mathcal{L}_{\text{instance}}  \label{final}
\end{equation}

\section{Experiments}

\begin{table}[t]
\setlength{\tabcolsep}{4pt}
\setlength{\extrarowheight}{-0.5pt}
\begin{center}
\begin{small}
\begin{tabular}{lcccc}
\toprule
\textbf{Method} & \textbf{Arch} & \begin{tabular}[c]{c}\textbf{Batch}\\ \textbf{size}\end{tabular} & \textbf{pre-training} & \textbf{ImageNet}  \\
 & & & \# epochs & Top-1 Acc  \\
\shline
Random \cite{goyal2019scaling} & R-50 &-& - & 13.7 \\
\midrule 
\multicolumn{5}{l}{\emph{Pre-training on ImageNet:}} \\
Supervised \cite{misra2020self} & R-50 &256& 90 & 75.9 \\
\shline
\multicolumn{5}{l}{\emph{Pre-training on MS COCO:}} \\ %
SimCLR \cite{chen2020simple} & \multirow{5}{*}{R-50} &512& 800 & 50.9  \\
MoCo v2 & &512& 800 & 55.1  \\
BYOL \cite{grill2020bootstrap} &  &512& 800 & 57.8  \\
ORL \cite{xie2021unsupervised} &  &512& 800 & 59.0  \\
UniVIP (ours) &  &512& 800 & \cellcolor{DnCBG}\textbf{60.2}    \\
\midrule 
\multicolumn{5}{l}{\emph{Pre-training on COCO+:}} \\
BYOL \cite{grill2020bootstrap} & \multirow{3}{*}{R-50} &512& 800 & 59.6  \\
ORL \cite{xie2021unsupervised}& &512& 800 & 60.7  \\
UniVIP (ours) & &512& 800 & \cellcolor{DnCBG}\textbf{63.0}  \\
\midrule 
\multicolumn{5}{l}{\emph{Pre-training on ImageNet:}} \\
InstDis \cite{wu2018unsupervised} & \multirow{11}{*}{R-50}&256& 200 & 58.5   \\
MoCo \cite{he2019momentum} & &256& 200 & 60.6   \\

CPC v2 \cite{lai2019contrastive} & &512& 200 & 63.8  \\
SimCLR \cite{chen2020simple} & &4096& 200 & 66.5   \\
MoCo v2 \cite{chen2020improved} & &256& 200 & 67.7  \\

SwAV \cite{caron2020unsupervised} & &4096& 200 & 69.1  \\
SimSiam \cite{chen2021exploring} & &256& 200 & 70.0 \\

InfoMin \cite{zhao2021what} & &256& 200 & 70.1  \\
BYOL \cite{grill2020bootstrap} & &4096& 200 & 70.6   \\

UniVIP (ours) & &4096& 200 & \cellcolor{DnCBG}\textbf{73.1}  \\
UniVIP (ours) & &4096& 300 & \cellcolor{DnCBG}\textbf{74.2}  \\
\bottomrule
\end{tabular}
\end{small}
\end{center}
\vspace{-4mm}
\caption{\textbf{Comparison of SOTA self-supervised learning methods.} 
UniVIP is pre-trained on non-iconic COCO(+), and single-centric-object ImageNet. The COCO(+) and ImageNet pre-trained UniVIP all achieve state-of-the-art performance than previous SSL methods. For evaluation, a linear classifier is trained on ImageNet, we report top-1 accuracy on the ImageNet \texttt{val} set.}
\vspace{-3mm}
\label{tabimage}
\end{table}

\begin{table}[]
\centering
\scalebox{0.85}{
\begin{tabular}{l|c|l  |l }
\shline
Method &\begin{tabular}[c]{c}Pre-train\\ data\end{tabular} & AP$^{bb}$ & AP$^{mk}$ \\ \shline
\gray{Rand Init}\cite{he2019momentum} &- &31.0 &28.5  \\
Supervised \cite{he2019momentum} & ImageNet  &38.9 &35.4   \\ \hline
SimCLR\cite{chen2020simple}  & COCO   &37.0\rbf{-1.9}  &33.7\rbf{-1.7}  \\ 
MoCov2\cite{chen2020improved}  & COCO   &38.5\rbf{-0.4}    &34.8\rbf{-0.6}  \\ 
BYOL\cite{grill2020bootstrap}  & COCO   &39.5\gbf{+0.6}  &35.6\gbf{+0.2}  \\ 
ORL\cite{xie2021unsupervised}  & COCO   &40.3\gbf{+1.4}  &36.3\gbf{+1.9} \\ 
Ours  & COCO   & \bf{40.8}\gbf{+1.9}  &\bf{36.8}\gbf{+1.4} \\ 
\hline
BYOL\cite{grill2020bootstrap}  & COCO+   &40.0\gbf{+1.1} &36.2\gbf{+0.8} \\ 
ORL\cite{xie2021unsupervised}  & COCO+   &40.6\gbf{+1.7}  &36.7\gbf{+1.3} \\ 
Ours  & COCO+   &\bf{41.1}\gbf{+2.2}  &\bf{37.1}\gbf{+1.7}\\ 
\hline
InsDis\cite{wu2018unsupervised}  & ImageNet   & 37.4\rbf{-1.5} &34.1\rbf{-1.3} \\
PIRL\cite{misra2020self}   & ImageNet     & 37.5\rbf{-1.4} &34.0\rbf{-1.4} \\ 
SwAV\cite{caron2020unsupervised}   & ImageNet    & 38.5\rbf{-0.4} &35.4\rbf{0.0}  \\ 
MoCo\cite{he2019momentum}  & ImageNet   &38.5\rbf{-0.4} &35.1\rbf{-0.3} \\ 
MoCov2\cite{chen2020improved}  & ImageNet   &38.9\rbf{0.0}   &35.5\gbf{+0.1} \\ 
BYOL\cite{grill2020bootstrap}  & ImageNet  &40.7\gbf{+1.8} &36.9\gbf{+1.5} \\ 
Ours  & ImageNet   &\bf{41.6}\gbf{+2.7} &\bf{37.6}\gbf{+2.2} \\ 
\shline
\end{tabular}}
\vspace{-2mm}
\caption{\small{
\textbf{Results of object detection and instance segmentation fine-tuned on COCO with $1\times$ schedule}. We adopt Mask R-CNN R50-FPN, and report the bounding box AP and mask AP on COCO \texttt{val2017}. The COCO(+) pre-trained methods are trained for 800 epochs while the ImageNet pre-trained methods are trained for 200 epochs. UniVIP outperforms all supervised and supervised conterparts. \green{Green} means increase and \gray{gray} means decrease.
}
}
\label{tab:coco_1x_c4fpn}
\vspace{-3mm}
\end{table}

\noindent\textbf{Datasets.} We first pre-train models on the COCO \texttt{train2017} set that contains $\sim$118k images. COCO contains more natural and diverse scenes in the wild. Then, we perform self-supervised learning on a larger ``COCO+'' dataset (COCO \texttt{train2017} set plus COCO \texttt{unlabeled2017} set) to verify whether our method can benefit from more unlabeled natural data. Finally, for verifying the unification of our method, we pre-train models on the single-centric-object ImageNet dataset. ImageNet \texttt{train} consists of $\sim$1.28 million training images.

\noindent\textbf{Network architecture.} Following recent un-/self-supervised methods \cite{he2019momentum,liu2020self,roh2021spatially,zhao2021self,xie2021self,xie2021unsupervised}, we adopt ResNet-50~\cite{he2016deep} as the default backbone. The two branches are slightly different, with one using a regular backbone network, a regular projection head, and a prediction head, and the other using the momentum network with a moving average of the parameters of the regular backbone network and the projection head. 

\noindent\textbf{Image augmentations.}  In pre-training, the scene-level data augmentation strategy follows \cite{grill2020bootstrap}. The setting of the scene image augmentation include: random horizontal flipping, random color jittering, random grayscale conversion, random Gaussian blurring, and solarization. For the instance-level augmentation, we directly crop the regions on the input images and resize them to 96$\times$96. The subsequent augmentations exactly follow the scene-level ones.

\noindent\textbf{Optimization.} For pre-training on the COCO(+), we fully follow the setting of \cite{xie2021unsupervised}. Specifically, we use the SGD optimizer with a weight decay of 0.0001 and a momentum of 0.9. We adopt the cosine learning rate decay schedule~\cite{loshchilov2016sgdr} with a base learning rate of 0.2, and the batch size is set to 512 by default. We train our models for 800 epochs with a warm-up period of 4 epochs. The exponential moving average parameter $m$ starts from 0.99 and is increased to 1 during training. Similarly, following the setting of BYOL \cite{grill2020bootstrap}, we pre-train the models on ImageNet.

\begin{table}[]
\centering
\scalebox{0.85}{
\begin{tabular}{l|c|l |l}
\shline
Method &\begin{tabular}[c]{c}Pre-train\\ data\end{tabular} & AP$^{bb}$ & AP$^{mk}$ \\ \shline
\gray{Rand Init}\cite{he2019momentum}&-  &36.7  &33.7  \\
Supervised\cite{he2019momentum} & ImageNet  &40.6&36.8   \\ \hline

BYOL\cite{grill2020bootstrap}  & COCO    &40.8\gbf{+0.2}  &37.0\gbf{+0.2}  \\ 
Ours  & COCO   &\bf{42.2}\gbf{+1.6}   &\bf{38.2}\gbf{+1.4}  \\ 
\hline
BYOL\cite{grill2020bootstrap}  & COCO+    &41.4\gbf{+0.8}  &37.4\gbf{+0.6} \\ 
Ours  & COCO+    &\bf{42.8}\gbf{+2.2}   &\bf{38.6}\gbf{+1.8}  \\ 
\hline

MoCo\cite{he2019momentum}   & ImageNet    &40.8\gbf{+0.2}  &36.9\gbf{+0.1} \\ 
MoCov2\cite{chen2020improved}  & ImageNet     &40.9\gbf{+0.3} &37.0\gbf{+0.2} \\

BYOL\cite{grill2020bootstrap}  & ImageNet   &42.2\gbf{+1.6} &38.0\gbf{+1.2} \\ 
Ours  & ImageNet    &\bf{43.1}\gbf{+2.5} &\bf{38.8}\gbf{+2.0} \\ 

\shline
\end{tabular}}
\vspace{-2mm}
\caption{\small{
\textbf{Results of object detection and instance segmentation fine-tuned on COCO with $2\times$ schedule}. 
The COCO(+) pre-trained methods are trained for 800 epochs while the ImageNet pre-trained methods are trained for 200 epochs. UniVIP achieves the state-of-the-art performance. 
}
}
\label{tab:coco_2x_c4fpn}
\vspace{-3mm}
\end{table}

\begin{table}[!t]
\centering
\scalebox{0.8}{
\begin{tabular}{l|c|c| ll }
\shline
Method & Epoch &\begin{tabular}[c]{c}Pre-train\\ data\end{tabular} & AP$^{bb}$ & AP$^{mk}$ \\ \shline
\gray{Rand Init} & - &- &31.0 &28.5  \\ 
Supervised                       & 90 & ImageNet&38.9           &35.4            \\ \hline
Self-EMD~\cite{liu2020self} & 800 & COCO &39.3\gbf{+0.4} &-  \\ 
DenseCL~\cite{wang2020dense} & 800 & COCO &39.6\gbf{+0.7} &35.7\gbf{+0.3}  \\ 
Resim-FPN~\cite{xiao2021region} & 200 & ImageNet &39.8\gbf{+0.9} &36.0\gbf{+0.6}  \\ 
DetCo~\cite{xie2021detco}       & 200& ImageNet&40.1\gbf{+1.2} &36.4\gbf{+1.0}  \\

DenseCL~\cite{wang2020dense} & 200 & ImageNet &40.3\gbf{+1.4} &36.4\gbf{+1.0}  \\ 
Resim-FPN~\cite{xiao2021region} & 400 & ImageNet &40.3\gbf{+1.4} &36.4\gbf{+1.0}  \\ 

BYOL~\cite{grill2020bootstrap} & 300 & ImageNet &40.9\gbf{+2.0} &37.0\gbf{+1.6} \\
SCRL~\cite{roh2021spatially}    & 200&ImageNet &41.0\gbf{+2.1} &37.5\gbf{+2.1}  \\ 
SCRL~\cite{roh2021spatially}    & 1000&ImageNet &41.4\gbf{+2.5} &37.9\gbf{+2.5}  \\ 
InsLoc-FPN~\cite{yang2021instance} &200 &ImageNet &41.4\gbf{+2.5} &37.1\gbf{+1.7}  \\ 
InsLoc-FPN~\cite{yang2021instance} &400 &ImageNet &42.0\gbf{+3.1} &37.6\gbf{+2.2}  \\ 
\hline
\multirow{4}*{UniVIP} & 800 & COCO &40.8\gbf{+1.9} &36.8\gbf{+1.4}  \\    
     & 800 & COCO+ &41.1\gbf{+2.2} &37.1\gbf{+1.7}  \\    
     & 200 & ImageNet &41.6\gbf{+2.8} &37.6\gbf{+2.1}  \\ 
     & 300 & ImageNet&\bf{42.2}\gbf{+3.3} &\bf{38.2}\gbf{+2.8}  \\ 
\shline
\end{tabular}}
\vspace{-2mm}
\caption{\textbf{Comparison with other self-supervised object detection methods.} We report the results with $1\times$ schedule on COCO. UniVIP-200ep is on par with previous state-of-the-art, and UniVIP-300ep achieves the best performance. 
}
\label{tab:ssldetec}
\vspace{-3mm}
\end{table}

\subsection{Linear evaluation on ImageNet}

We compare our method with other prevailing algorithms pre-trained on COCO, COCO+, and ImageNet datasets in Table~\ref{tabimage}. All these methods share the same backbone for fair comparison and evaluation by linear probing. For COCO dataset, our model achieves 60.2\% top-1 accuracy with linear probing. It outperforms the previous best algorithm ORL by 1.2\% at the same training epochs, and even approaches the performance of ORL with a much larger dataset (60.7\% pre-trained on COCO+). Our algorithm relieves the need for a larger training dataset for self-supervised learning, and is able to obtain a decent result (60.2\%) with only $\sim$118k images.
Notably, our method achieves 63.0\% top-1 accuracy and outperforms ORL by 2.3\% when applied to COCO+, which means UniVIP becomes better using a larger dataset. It should be emphasized that UniVIP is a unified framework, which can also be applied with the single-centric-object dataset. Here we use the popular ImageNet as an example. With the same pre-training epochs, UniVIP outperforms BYOL by 2.5\%, which is a state-of-the-art self-supervised learning method designed delicately for the single-centric-object dataset. These results indicate that UniVIP is a unified method, which can be pre-trained with any images for visual self-supervised representation learning.

\subsection{Object detection and segmentation}

\textbf{COCO with 1$\times$ and 2$\times$ schedule.} We perform object detection and segmentation experiments using Mask R-CNN detector \cite{he2017mask} with R50-FPN \cite{lin2017feature} implemented in Detectron2 \cite{wu2019detectron2}. We fine-tune all layers end-to-end on COCO \texttt{train2017} set and evaluate on \texttt{val2017} ($\sim$5k images). The schedule is the default 1$\times$ or 2$\times$ following the same setup in \cite{he2019momentum,zhao2021what}. In Table \ref{tab:coco_1x_c4fpn}, we show the results of the learned representation by different self-supervised methods pre-training on different datasets. For fair comparison, all these methods are pre-trained with the same epochs. It can be observed that our method achieves the best results with $40.8\%$ bbox mAP ($\text{AP}^{bb}$) and $36.8\%$ mask mAP ($\text{AP}^{mk}$) when pre-trained on COCO. It outperforms the ImageNet supervised counterpart by 1.9\% and 1.4\%, and BYOL results by 1.3\% and 1.2\%. Similarly, the COCO+ pre-trained UniVIP reaches $41.1\%$ $\text{AP}^{bb}$ and $37.1\%$ $\text{AP}^{mk}$, which surpasses the supervised by 2.2\% and 1.7\%, and BYOL results by 1.1\% and 0.9\%. Notably, the results also show excellent performance when pre-trained on single-centric-object ImageNet. Our method yields 3.3\% $\text{AP}^{bb}$ and 2.8\% $\text{AP}^{mk}$ improvements over the supervised. Simultaneously, as shown in Table \ref{tab:coco_2x_c4fpn}, the impressive performance is still retained when the pre-trained models are trained with longer epochs in downstream dense prediction tasks.

\textbf{Comparison with current self-supervised object detection methods.} It should be emphasized that our method also outperforms current state-of-the-art self-supervised object detection methods \cite{liu2020self,wang2020dense,xiao2021region,yang2021instance,xie2021detco,roh2021spatially} as shown in Table \ref{tab:ssldetec}, even though part methods \cite{xiao2021region,yang2021instance} pre-training with FPN \cite{lin2017feature}. Meanwhile, UniVIP acquires more gains with longer pre-training epochs, which shows the great potential of our framework. These phenomena indicate that learning the versatile visual representation can effectively enhance the transfer ability of models. Although our method requires about 35\% computation than the BYOL, but the performance of 300-epoch UniVIP outperforms the 1000-epoch SCRL, which is an object detection self-supervised method based on BYOL. The results show that the computation of UniVIP brings a large gain. 

\textbf{COCO with longer training iterations.} Moreover, in \cite{he2019rethinking}, it discusses that ImageNet pre-training shows rapid convergence than random initialization at the early stage of training but the final performance is not better than the model trained from scratch. As shown in Table \ref{lr_schedule}, the scratch can narrow the gap and even surpass BYOL at last, even though the BYOL pre-training provides a slightly better initial point. Notably, UniVIP goes beyond the limitation and the noticeable gain is preserved even in longer schedules. We argue that UniVIP provides versatile representations of quality that the previous pre-training methods have not yet achieved.

\begin{table}
\small
\centering
\scalebox{0.85}{
\begin{tabular}{c|ccc}
\multirow{2}*{pretrain} & \multicolumn{3}{c}{LR schedule} \\
\cline{2-4} & $1\times$ & $2\times$ & $6\times$ \\ \hline \hline
Random \cite{zhao2021what} & 31.0 & 36.7 & 42.7 \\
Supervised \cite{zhao2021what} & 38.9\gbf{+7.9} & 40.6\gbf{+3.9} & 42.6\rbf{-0.1} \\
BYOL \cite{grill2020bootstrap} & 40.9\gbf{+9.9} & 42.3\gbf{+5.6} & 42.6\rbf{-0.1} \\
Ours & \textbf{42.2}\gbf{+11.2} & \textbf{43.5}\gbf{+6.8} & \textbf{44.0}\gbf{+1.3}  \\ \hline
\end{tabular}
}
\vspace{-2mm}
\caption{\textbf{Results of object detection fine-tuned on COCO with longer training iterations.} The results of UniVIP-300ep object detection $\text{AP}^{bb}$ on COCO \texttt{val2017} with training schedules from $1\times$ ($90$k iterations) to $6\times$ ($540$k iterations).
}
\label{lr_schedule}
\vspace{-2mm}
\end{table}

\begin{table}
\centering
\small

\scalebox{0.85}{

\begin{tabular}{lccccc}
\toprule
\multirow{2}{*}{Method} & \multirow{2}{*}{\begin{tabular}[c]{c}Pre-train\\ data\end{tabular}} & \multicolumn{2}{c}{1\% labels} & \multicolumn{2}{c}{10\% labels} \\
          &          & Top-1 & Top-5 & Top-1 & Top-5 \\ \midrule
Random~\cite{xie2021unsupervised}    & -        & 1.6      & 5.0      & 21.8      & 44.2      \\
Supervised~\cite{zhai2019s4l} & ImageNet & 25.4      & 48.4      &  56.4     & 80.4      \\ \midrule
SimCLR~\cite{chen2020simple}     & COCO     & 23.4      & 46.4      & 52.2      & 77.4      \\
MoCo v2~\cite{chen2020improved}    & COCO     & 28.2      & 54.7      & 57.1      & 81.7      \\
BYOL~\cite{grill2020bootstrap}       & COCO     & 28.4      & 55.9      & 58.4      & 82.7      \\
ORL~\cite{xie2021unsupervised}        & COCO     & 31.0      & 58.9      & 60.5      & 84.2     \\ 
Ours        & COCO     & \textbf{31.6}      & \textbf{59.7}      & \textbf{61.3}      & \textbf{85.6}      \\ 
\midrule
BYOL~\cite{grill2020bootstrap}       & COCO+    & 28.3      & 56.0      & 59.4      & 83.6      \\
ORL~\cite{xie2021unsupervised}      & COCO+    & 31.8      & 60.1      &  60.9     & 84.4      \\
Ours      & COCO+    & \textbf{32.3}      & \textbf{60.9}      &  \textbf{61.7}     & \textbf{85.7}      \\
\midrule
BYOL~\cite{grill2020bootstrap}       & ImageNet   & 51.5      & 76.7      & 64.7      & 86.6      \\
Ours      & ImageNet    & \textbf{53.0}      & \textbf{78.8}      &  \textbf{67.1}     & \textbf{88.5}      \\

\bottomrule
\end{tabular}
}
\hspace{0.01\textwidth}
\vspace{-0.2cm}
\caption{\textbf{Semi-supervised learning on ImageNet.} The COCO(+) pre-trained methods are trained for 800 epochs while the ImageNet pre-trained methods are trained for 200 epochs. We fine-tune all models with 1\% and 10\% ImageNet labels, and report both top-1 and top-5 center-crop accuracy on the ImageNet \texttt{val} set.}
\label{tab:semi}
\vspace{-0.4cm}
\end{table}

\subsection{Semi-supervised learning}

We evaluate the performance obtained when fine-tuning UniVIP’s representation on the semi-supervised learning task, following the protocol of \cite{xie2021unsupervised,grill2020bootstrap}. In particular, we acquire 1\% and 10\% labeled data from ImageNet's \texttt{train} set. Furthermore, we fine-tune our models on these two training subsets and report both top-1 and top-5 accuracies on the \texttt{val} set of ImageNet in Table \ref{tab:semi}. UniVIP consistently outperforms previous SOTA methods when pre-trained on the non-iconic and single-centric-object dataset.

\subsection{Ablation studies}
\label{ablation}

\textbf{Effect of different levels.} As shown in Table \ref{tab:ab1}, it shows the effectiveness of our proposed scene, scene-instance, and instance levels. The results of Table \ref{tab:ab1}(a) are that we reproduce the performance of BYOL pre-training on MS COCO as the baseline. Compared with the baseline, the scene-level pre-training (Table \ref{tab:ab1}(b)) can improve the performance of the neural network. Moreover, adding either the scene-instance-level (Table \ref{tab:ab1}(c)) or instance-level pre-training (Table \ref{tab:ab1}(d)) based on the scene level can enhance the performance, meanwhile, the best results (Table \ref{tab:ab1}(e)) are acquired by adding three terms.

\begin{table}
\centering
\scalebox{0.825}{
\begin{tabular}{c|c|c|c|l|l}
\shline
    & Scene & Scene-instance & Instance & Top-1 & mAP \\ \shline
(a) & $\times$ & $\times$ & $\times$ & 57.3 & 39.5 \\ \shline
(b) & \checkmark & $\times$ & $\times$ & 57.7\gbf{+0.4} & 39.6\gbf{+0.1} \\ \hline
(c) & \checkmark & \checkmark  & $\times$ & 58.1\gbf{+0.8} & 40.4\gbf{+0.9} \\ \hline
(d) & \checkmark & $\times$ & \checkmark & 59.6\gbf{+2.3} & 40.5\gbf{+1.0} \\ \hline
(e) & \checkmark & \checkmark & \checkmark & \textbf{60.2}\gbf{+2.9} &  \textbf{40.8}\gbf{+1.3} \\ \shline
\end{tabular}
}
\vspace{-2mm}
\caption{{\bf Ablations for UniVIP: Effect of different levels when pre-trained on COCO dataset.} We report linear evaluation on ImageNet and detection results on COCO.}
\label{tab:ab1}
\vspace{-4mm}
\end{table}

\textbf{Effect of the scene similarity.} Table \ref{tab:combination}(a) ablates the effect of the scene similarity. The ``no'' uses the same methods with UniVIP except for adopting two random scene views, which results in scene inconsistency. UniVIP obviously outperforms its results. Therefore, the scene similarity is crucial to self-supervised learning in natural images.

\textbf{Effect of region candidates.} For validating the effect of region candidates, Table \ref{tab:combination}(b) shows the results of the baseline in ``none''. Then, we observe that the ``ground truth'' of COCO can develop the performance of the pre-trained model, validating that the gains are truly due to the instance-level representation learning mechanism. Furthermore, the performance of the ``naive'' strategy is slightly better than the ground truth. In fact, COCO only contains manual annotations for 80 classes but the scene images have extensive unknown classes. The naive method can obtain regions where more categories are located, even if they are very crude. In this case, the diversity can make up for the inaccuracy. Nevertheless, this phenomenon does not mean that the regions containing the instance are useless. The ``selective search'' results indicate that the instance-based regions effectively improve the performance. The more diverse region proposal method can greatly enhance the feasibility of the model, and also evaluate that the gains are truly due to instance-level representation pre-training.

\textbf{Effect of the number of instance-based views.} As shown in Table \ref{tab:combination}(c), we observe that UniVIP already outperforms the state-of-the-art method \cite{xie2021unsupervised} when the $K$ is set as 2. The best results can be obtained when further increasing $K$ to 4. Meanwhile, the performance slightly degrades as $K$ increases. We argue that using 4 candidate proposals in the overlapping regions can meet most scenes, and more candidates may bring noises, thus hurting the performance.

\begin{table}
\begin{center}
\scalebox{1.0}{
\begin{tabular}{c|c|c|c}
\shline
 \multirow{3}*{(a)} & \textbf{Scene similarity}  & Top-1 & mAP \\
\cline{2-4}
 & no & 59.3 & 40.3 \\
 & yes & \textbf{60.2} & \textbf{40.8} \\
\shline
\multirow{5}*{(b)} 
 &  \textbf{Region candidates} & Top-1 & mAP \\
\cline{2-4}
 & none & 57.7 & 39.6 \\
 & ground truth & 58.5 & 40.3 \\
 & naive & 58.9 & 40.5 \\
 & selective search & \textbf{60.2} & \textbf{40.8} \\
\shline
\multirow{5}*{(c)}
& \textbf{Number of instance views} & Top-1 & mAP \\
\cline{2-4}
 & 0 & 57.7 & 39.6 \\
 & 2 & 59.8 & 40.4 \\
 & 4 & \textbf{60.2} & \textbf{40.8} \\
 & 8 & 59.7 & 40.6 \\
\shline
\end{tabular}
}
\end{center}
\vspace{-5mm}
\caption{{\bf Ablations for UniVIP.} (a) Effect of the scene similarity. (b) Effect of region candidates. (c) Effect of the number of instance-based views. We report linear evaluation on ImageNet and detection results on COCO.}
\label{tab:combination}
\vspace{-5mm}
\end{table}

\section{Conclusion}
In this paper, we analyze the two problems of current visual self-supervised learning: 1) The SSL methods with semantic consistency assumption would be infeasible for non-iconic datasets since the random views of non-iconic images are semantic inconsistency. 2) The non-iconic SSL methods hardly extract versatile visual representations. To overcome the above problems, we introduce a novel unified self-supervised learning method called UniVIP. By simultaneously leveraging the similarity of scene-scene, the correlation of scene-instance, and the discrimination of instance-instance, UniVIP can improve the versatile performance of self-supervised learning with any images. The proposed UniVIP shows good versatility and scalability in multiple downstream visual tasks, such as image classification, semi-supervised learning, object detection and segmentation. We expect that our study can attract the community’s attention to
more versatile un-/self-supervised representation learning from natural images.

\noindent \textbf{Limitations.} In this work, we validate the performance of UniVIP by constructing experiments on COCO(+) dataset, and further scale our method up on ImageNet dataset. Compared with previous SSL methods, the COCO(+) and ImageNet pre-trained models all perform impressive improvements. In future, we will scale UniVIP with larger architectures and datasets \cite{zhou2017places,thomee2016yfcc100m,sun2017revisiting} to unleash its potential.

\noindent \textbf{Acknowledgement.} This work was supported by Key-Area Research and Development Program of Guangdong Province (No.2021B0101410003), National Natural Science Foundation of China under Grants No.62002357, No.62176254, No.61976210, No.61876086,  No.62076235 and No.62006230, and the IAF-ICP Funding Initiative.

{\small
\bibliographystyle{ieee_fullname}
\bibliography{egbib}
}

\clearpage
\begin{center}
{\Large\bf{Appendix}\\
}
\end{center}
\maketitle
\appendix

\section{Implementation details}
\subsection{Linear probing}
Following common practice, we evaluate the representation quality by linear probing. After self-supervised pre-training, we remove the MLP heads and train a supervised linear classifier on frozen features. We use the SGD optimizer, and the setting of batch size, weight decay, and learning rate depends on the type of dataset. Since the linear probing strategies of COCO(+) and ImageNet are different in previous methods \cite{xie2021unsupervised,grill2020bootstrap}, we evaluate the performance of COCO(+) pre-trained models by fully following ORL \cite{xie2021unsupervised} while we validate the linear classifier of ImageNet pre-trained models strictly following BYOL \cite{grill2020bootstrap}. We evaluate single-crop top-1 accuracy in the validation set.


\subsection{How to create overlapping Regions containing multi-instance}

\label{create}

We create two scene views $s_1$, $s_2$ with the overlapping regions containing $K$ identical objects as Algorithm \ref{alg:Algorithm1}, several images cannot meet the requirements of K candidates, and we adopt a random strategy to generate supplementary boxes. That's setting the minimum scale to 64 pixels,the range of aspect ratio is between 1/3 and 3/1, and the maximum IoU threshold is 0.5. 

\textbf{Effect of the iterations of creating overlapping regions.} Table \ref{tab:ab2} ablates the effect of the iterations of creating overlapping regions. It can be observed that the pre-training algorithm is robust to the hyper-parameters $iters$, thus we set the iterations $iters$ as 20 by default.

\begin{table}[ht]
\begin{center}
\begin{tabular}{c|c|c|c}
\hline
Method & Iterations & Top-1 & mAP \\
\hline
\multirow{3}*{UniVIP} 
 & 10 & 60.0 & 40.7 \\
 & 20 & 60.2 & 40.8 \\
 & 30 & 60.2 & 40.8 \\
\hline
\end{tabular}
\end{center}
\caption{{\bf Ablations for UniVIP: Effect of the iterations of creating overlapping area when pre-trained on MS COCO dataset.} We report linear evaluation on ImageNet and detection result on MS COCO.}
\label{tab:ab2}
\end{table}

\begin{algorithm}[t]
\caption{Create Overlapping Regions}
\label{alg:Algorithm1}
\hspace*{0.02in} {\bf Input:}\\
\hspace*{0.2in} $A$ is an input image. \\
\hspace*{0.2in} $T = [x,y,h,w]$ is the coordinates of overlapping areas. \\
\hspace*{0.2in} $boxes$ is the set of object coordinates for each image. \\
\hspace*{0.2in} $K$ is the number of instances in the created overlapping regions. \\
\hspace*{0.2in} $iters$ is the iterations of creating overlapping regions containing multiple instances. \\
\hspace*{0.02in} {\bf Output:} \\
\hspace*{0.2in} $s_1$ and $s_2$ are the scene views of the input image.
\begin{algorithmic}[1]

\State obtain the object-based region by selective search for each image $A$, and filter some redundancy to get $boxes$,\\
set $i$ as 0, \\
create overlapping areas during random cropping, then get the two scene views $s_1$, $s_2$ and the coordinates $T$ of overlapping regions, \\
set $j$ as 0, \\
\textbf{if} $i$ $\le$ $iters$: \\
\hspace*{0.2in} \textbf{for} box in $boxes$ \textbf{do}:   \\
\hspace*{0.4in} \textbf{if} box[0] $\leq$ $x$ \textbf{and} box[1] $\leq$ $y$ \textbf{and} box[0] + box[2] $\geq$ $x$ + $h$ \textbf{and} box[1] + box[3] $\geq$ $y$ + $w$ \textbf{do}: \\
\hspace*{0.6in} $j$ = $j$ + 1 \\
\hspace*{0.6in} \textbf{if} $j$ == $K$   \textbf{do}: \\
\hspace*{0.8in} \textbf{return} $s_1$, $s_2$ \\
\hspace*{0.2in} $i$ = $i$ + 1 \\
\hspace*{0.2in} back to the step 3 \\
\textbf{else}: \\
\hspace*{0.2in} ensure the overlapping regions have K regions, then \textbf{return} $s_1$, $s_2$.
\end{algorithmic}
\end{algorithm}
\subsection{How to filter the certain number of redundancy of generated proposals}
According to the subsection 3.2 of this paper, we filter the certain number of redundancy of generated proposals. The strategy also follows Section \ref{create}, includes: the minimal scale as $64$ pixels, the range of aspect ratio between $1/3$ and $3/1$, and the maximal IoU among the object-based regions as $0.5$.

\subsection{How to solve the optimal transport problem}

\begin{table*}[!ht]
\centering
\scalebox{1.15}{
\begin{tabular}{l|c|l l l |l l l}
\shline
\multirow{2}{*}{Method} &\multirow{2}{*}{\begin{tabular}[c]{c}Pre-train\\ data\end{tabular}} & \multicolumn{6}{c}{Mask R-CNN R50-C4 COCO 1$\times$}  \\ \cline{3-8} 
 & & AP$^{bb}$ & AP$^{bb}_{50}$ & AP$^{bb}_{75}$ & AP$^{mk}$ & AP$^{mk}_{50}$ & AP$^{mk}_{75}$  \\ \shline
\gray{Rand Init}\cite{he2019momentum} &- &26.4 &44.0 &27.8 &29.3 &46.9 &30.8  \\
Supervised \cite{he2019momentum} & ImageNet  &38.2 &58.2 &41.2 &33.3 &54.7 &35.2   \\
DetCo \cite{xie2021detco}    & ImageNet &39.8\gbf{+1.6} &59.7\gbf{+1.5} &43.0\gbf{+1.8} &34.7\gbf{+1.4} &56.3\gbf{+1.6} &36.7\gbf{+1.5}
\\ \hline
SimCLR\cite{chen2020simple}  & COCO   &34.4\rbf{-3.8} &54.0\rbf{-4.2} &36.4\rbf{-4.8} &30.7\rbf{-2.6} &50.6\rbf{-4.1} &32.6\rbf{-2.6}  \\ 
MoCov2\cite{chen2020improved}  & COCO   &37.6\rbf{-0.6} &57.0\rbf{-1.2} &40.4\rbf{-0.8} &33.0\rbf{-0.3}&53.8\rbf{-0.9} &34.9\rbf{-0.3}  \\ 
BYOL\cite{grill2020bootstrap}  & COCO   &38.1\rbf{-0.1} &57.4\rbf{-0.8} &40.5\rbf{-0.7} &33.5\gbf{+0.2} &54.2\rbf{-0.5} &35.5\gbf{+0.3}  \\ 
Ours  & COCO   &39.3\gbf{+1.1} &58.9\gbf{+0.7} &42.2\gbf{+1.0} &34.4\gbf{+1.1} &55.7\gbf{+1.0} &36.5\gbf{+1.3}  \\ 
\hline
BYOL\cite{grill2020bootstrap}  & COCO+   &38.8\gbf{+0.6} &58.8\gbf{+0.6} &42.0\gbf{+0.8} &34.1\gbf{+0.8} &55.5\gbf{+0.8} &36.5\gbf{+1.3}  \\ 
Ours  & COCO+   &\bf 40.1\gbf{+1.9} &\bf 60.0\gbf{+1.8} &\bf 43.2\gbf{+2.0} &\bf 35.2\gbf{+1.9} &\bf 56.7\gbf{+2.0} &\bf 37.6\gbf{+1.4}  \\ 
\shline
\end{tabular}}
\caption{\small{
\textbf{Results of object detection and instance segmentation fine-tuned on COCO with $1\times$ schedule}. We adopt Mask R-CNN R50-C4, and report the bounding box AP and mask AP on COCO \texttt{val2017}.
The COCO(+) pre-trained methods are trained for 800 epochs. UniVIP outperforms all supervised and unsupervised counterparts. 
}
}
\label{tab:coco_1x_c4}
\end{table*}

\begin{table*}
\centering
\scalebox{1.15}{
\begin{tabular}{l|c|l l l |l l l}
\shline
\multirow{2}{*}{Method} &\multirow{2}{*}{\begin{tabular}[c]{c}Pre-train\\ data\end{tabular}} & \multicolumn{6}{c}{Mask R-CNN R50-C4 COCO 2$\times$}  \\ \cline{3-8} 
 & & AP$^{bb}$ & AP$^{bb}_{50}$ & AP$^{bb}_{75}$ & AP$^{mk}$ & AP$^{mk}_{50}$ & AP$^{mk}_{75}$  \\ \shline
\gray{Rand Init}\cite{he2019momentum} &- &35.6 &54.6 &38.2 &31.4 &51.5 &33.5  \\
Supervised \cite{he2019momentum} & ImageNet  &40.0 &59.9 &43.1 &34.7 &56.5 &36.9   \\ 
DetCo\cite{xie2021detco}  & ImageNet  &41.3\gbf{+1.3} &61.2\gbf{+1.3} &45.0\gbf{+1.9} &35.8\gbf{+1.1} &57.9\gbf{+1.4} &38.2\gbf{+1.3} \\
\hline
BYOL\cite{grill2020bootstrap}  & COCO   &39.0\rbf{-1.0} &58.6\rbf{-1.3} &42.1\rbf{-1.0} &34.0\rbf{-0.7} &55.0\rbf{-1.5} &36.1\rbf{-0.8}  \\ 
Ours  & COCO   &41.0\gbf{+1.0} &60.6\gbf{+0.7} &44.6\gbf{+1.5} &35.6\gbf{+0.9} &57.4\gbf{+0.9} &37.8\gbf{+0.9}  \\ 
\hline
BYOL\cite{grill2020bootstrap}  & COCO+   &40.6\gbf{+0.6} &60.3\gbf{+0.4} &44.0\gbf{+0.9} &35.4\gbf{+0.7} &57.2\gbf{+0.7} &37.7\gbf{+0.8}  \\ 
Ours  & COCO+   &\bf 41.5\gbf{+1.5} &\bf 61.3\gbf{+1.4} &\bf 45.0\gbf{+1.9} &\bf 36.3\gbf{+1.6} &\bf 58.0\gbf{+1.5} &\bf 38.7\gbf{+1.8}  \\ 
\shline
\end{tabular}}
\caption{\small{
\textbf{Results of object detection and instance segmentation fine-tuned on COCO with $2\times$ schedule}. 
The COCO(+) pre-trained methods are trained for 800 epochs. UniVIP outperforms all supervised and unsupervised counterparts. 
}
}
\label{tab:coco_2x_c4}
\end{table*}

\begin{table*}
\centering
\scalebox{1.15}{
\begin{tabular}{l|c|c|l l l}
\shline
Method  &\begin{tabular}[c]{c}Pre-train\\ data\end{tabular}& Epoch & AP & AP$_{50}$ & AP$_{75}$ \\ \shline
\gray{Rand Init} & - &-&33.8 &60.2 &33.1 \\ 
Supervised                       &ImageNet& 90  &53.5           &81.3           &58.8 \\ \hline
InsDis~\cite{wu2018unsupervised} &ImageNet& 200 &55.2\gbf{+1.7} &80.9\rbf{-0.4} &61.2\gbf{+2.4} \\ 
PIRL~\cite{misra2020self}                 &ImageNet& 200 &55.5\gbf{+2.0} &81.0\rbf{-0.3} &61.3\gbf{+2.5} \\ 
SwAV~\cite{caron2020unsupervised}                 &ImageNet& 800 &56.1\gbf{+2.6} &82.6\gbf{+1.3} &62.7\gbf{+3.9} \\ 
MoCo~\cite{he2019momentum}                 &ImageNet& 200 &55.9\gbf{+2.4} &81.5\gbf{+0.2} &62.6\gbf{+3.8} \\ 
MoCov2~\cite{chen2020improved}             &ImageNet& 800 &57.4\gbf{+3.9} &82.5\gbf{+1.2} &64.0\gbf{+5.2} \\ 
DetCo \cite{xie2021detco}                     &ImageNet& 200 &57.8\gbf{+4.3} &82.6\gbf{+1.3} &64.2\gbf{+5.4} \\ 
DetCo  \cite{xie2021detco}                          &ImageNet& 800 &58.2\gbf{+4.7} &82.7\gbf{+1.4} &65.0\gbf{+6.2} \\ \hline
BYOL \cite{grill2020bootstrap} &COCO& 800 & 53.8\gbf{+0.3} & 79.9\rbf{-1.5} & 59.1\gbf{+0.3} \\
UniVIP            &COCO& 800 &56.5\gbf{+3.0} &82.3\gbf{+1.0} &62.6\gbf{+3.8} \\ \hline
BYOL \cite{grill2020bootstrap} &COCO+& 800 & 56.4\gbf{+2.9} &81.9\gbf{+0.6} &62.6\gbf{+3.8} \\
UniVIP        &COCO+& 800 &\bf 58.2\gbf{+4.7} &\bf 83.3\gbf{+2.0} &\bf 65.2\gbf{+6.4} \\ \shline
\end{tabular}}
\caption{\textbf{Object detection finetuned on PASCAL VOC07+12 using Faster RCNN with R50-C4.} UniVIP-800ep achieves the best performance when pre-trained on COCO+.
}
\label{tab:voc}
\end{table*}

\begin{table*}
\centering
\scalebox{1.3}{
\begin{tabular}{l|c|l l l}
\shline
\multirow{2}{*}{Method} &\multirow{2}{*}{\begin{tabular}[c]{c}Pre-train\\ data\end{tabular}} &\multicolumn{3}{c}{RetinaNet R50 1$\times$}  \\ \cline{3-5} 
& &AP &AP$_{50}$ &AP$_{75}$  \\ \shline
\gray{Rand Init}       &-     &24.5           &39.0           &25.7            \\ 
Supervised        &ImageNet   &37.4           &56.5           &39.7            \\ \hline
InsDis\cite{wu2018unsupervised} &ImageNet &35.5\rbf{-1.9} &54.1\rbf{-2.4} &38.2\rbf{-1.5}  \\ 
PIRL\cite{misra2020self}        &ImageNet &35.7\rbf{-1.7} &54.2\rbf{-2.3} &38.4\rbf{-1.3}  \\ 
SwAV\cite{caron2020unsupervised}   &ImageNet      &35.2\rbf{-2.2} &54.9\rbf{-1.6} &37.5\rbf{-2.2} \\ 
MoCo\cite{he2019momentum}    &ImageNet  &36.3\rbf{-1.1} &55.0\rbf{-1.5} &39.0\rbf{-0.7}  \\ 
MoCov2\cite{xie2021detco} &ImageNet &37.2\rbf{-0.2} &56.2\rbf{-0.3} &39.6\rbf{-0.1}  \\ 
DetCo\cite{xie2021detco}      &ImageNet     &38.4\gbf{+1.0} &57.8\gbf{+1.3} &\bf 41.2\gbf{+1.5}  \\ \hline
BYOL \cite{grill2020bootstrap}     &COCO     &36.0\rbf{-1.4} &54.5\rbf{-2.0} &38.5\rbf{-1.2}  \\ 
UniVIP     &COCO     &38.0\gbf{+0.6} &57.4\gbf{+0.9} &40.6\gbf{+0.9}  \\ 
\hline
BYOL \cite{grill2020bootstrap}     &COCO+     &37.0\rbf{-0.4} &56.2\rbf{-0.3} &39.5\rbf{-0.2}  \\ 
UniVIP     &COCO+     &\bf{38.5}\gbf{+1.1} &\bf{58.0}\gbf{+1.5} &41.0\gbf{+1.3}  \\ 
\shline
\end{tabular}}
\caption{\small{
\textbf{One-stage object detection fine-tuned on COCO}. 
The ImageNet pre-trained methods are trained for 200 epochs while the COCO(+) pre-trained methods  are trained for 800 epochs.
UniVIP outperforms all supervised and unsupervised counterparts. 
}
}
\label{tab:retina}
\end{table*}

In this paper, we adopt Sinkhorn Iteration algorithm \cite{cuturi2013sinkhorn} to solve the discrimination of instance-instance. 
The OT problem can be addressed by a fast iterative solution, which converts the optimization target into a non-linear but convex form with an entropic regularization term added. It is noted that the solution is not our contributions and belongs to textbook knowledge, and the details can refer to preliminary work \cite{cuturi2013sinkhorn,ge2021ota}.

\section{Compared with multi-crop BYOL}

In prior work \cite{xie2021unsupervised}, the results of the COCO pre-trained ORL show that simply adding more crops tends to hurt the performance since the operation will further intensify the inconsistent noise on non-iconic images. Meanwhile, another prior work \cite{caron2021emerging} shows that the multi-crop BYOL is also inferior to its baseline. These papers already indicate that multi-crop BYOL hurts the performance of models when pre-trained on either single-centric-object or non-iconic dataset, thus we do not construct the experiment of multi-crop BYOL.

\section{More experiment results}
All training settings of the below experiments strictly follow \cite{he2019momentum,xie2021detco}. We report object detection results on PASCAL VOC \cite{everingham2010pascal} and COCO  \cite{lin2014microsoft} datasets.

\textbf{Results of Mask R-CNN with R50-C4 using 1$\times$ and 2$\times$ schedule.} We perform object detection and segmentation experiments using Mask R-CNN detector \cite{he2017mask} with R50-C4 \cite{lin2017feature} implemented in Detectron2 \cite{wu2019detectron2}. We fine-tune all layers end-to-end on COCO \texttt{train2017} set and evaluate on \texttt{val2017} ($\sim$5k images). The schedule is the default 1$\times$ or 2$\times$ following the same setup in \cite{he2019momentum,zhao2021what}.
As shown in Table \ref{tab:coco_1x_c4} and Table \ref{tab:coco_2x_c4}, our UniVIP show impressive performance. The COCO+ pre-trained UniVIP outperforms all supervised and unsupervised counterparts, even though its pre-train data is only $\sim$241k images.
It should be emphasized that UniVIP surpasses the self-supervised object detection method DetCo \cite{xie2021detco}, although DetCo uses $\sim$1.28 million images to pre-train models.

\textbf{Results of PASCAL VOC07+12.} We also construct the experiments of object detection fine-tuned on PASCAL VOC07+12 using Faster RCNN with R50-C4 \cite{ren2015faster} in Table \ref{tab:voc}, fully following the setting of \cite{xie2021detco}. It can be observed that UniVIP is still better than DetCo, even the number of pre-train images and iterations is far less than DetCo.

\textbf{Results of RetinaNet.} In Table \ref{tab:retina}, we compared with other methods on one-stage object detection method RetinaNet \cite{lin2020focal}, and the setting of experiments fully following \cite{xie2021detco}. Notably, our UniVIP also surpasses the DetCo.

The above experiments demonstrate the effectiveness and potential of our UniVIP.


\end{document}